\documentclass[10pt,twocolumn,letterpaper]{article}

\usepackage{iccv}
\usepackage{times}
\usepackage{epsfig}
\usepackage{graphicx}
\usepackage{amsmath}
\usepackage{amssymb}

\usepackage{wrapfig}
\usepackage{multirow}
\usepackage[normalem]{ulem}
\useunder{\uline}{\ul}{}
\usepackage{xr}
\usepackage{amsmath}
\usepackage{amssymb}
\usepackage{booktabs}

\newcommand{\chair}[0]{\emph{chair}}
\newcommand{\lamp}[0]{\emph{lamp}}
\newcommand{\bench}[0]{\emph{bench}}
\usepackage[table,xcdraw]{xcolor}

\usepackage[pagebackref=true,breaklinks=true,letterpaper=true,colorlinks,bookmarks=false]{hyperref}

\usepackage[capitalize]{cleveref}
\crefname{section}{Sec.}{Secs.}
\Crefname{section}{Section}{Sections}
\Crefname{table}{Table}{Tables}
\crefname{table}{Tab.}{Tabs.}

\iccvfinalcopy % *** Uncomment this line for the final submission

\def\iccvPaperID{****} % *** Enter the ICCV Paper ID here

\ificcvfinal\fi

\begin{document}

%%%%%%%%% TITLE
\title{Fine-Tuned but Zero-Shot 3D Shape Sketch View Similarity and Retrieval}

% \author[1,2]{Gianluca Berardi}
% \author[2]{Yulia Gryaditskaya}
% \affil[1]{Department of Computer Science and Engineering (DISI), University of Bologna, Italy}
% \affil[2]{CVSSP and Surrey Institute for People-Centred AI, University of Surrey, UK}

\author{Gianluca Berardi$^{1,2}$\\
\and
Yulia Gryaditskaya$^{2}$
\and
$^{1}$Department of Computer Science and Engineering (DISI), University of Bologna, Italy\\
$^{2}$CVSSP and Surrey Institute for People-Centred AI, University of Surrey, UK\\
}

\maketitle
% Remove page # from the first page of camera-ready.
\ificcvfinal\thispagestyle{empty}\fi

%%%%%%%%% ABSTRACT
\begin{abstract}
Recently, encoders like ViT (vision transformer) and ResNet have been trained on vast datasets and utilized as perceptual metrics for comparing sketches and images, as well as multi-domain encoders in a zero-shot setting. However, there has been limited effort to quantify the granularity of these encoders.
Our work addresses this gap by focusing on multi-modal 2D projections of individual 3D instances. This task holds crucial implications for retrieval and sketch-based modeling. We show that in a zero-shot setting, the more abstract the sketch, the higher the likelihood of incorrect image matches. Even within the same sketch domain, sketches of the same object drawn in different styles, for example by distinct individuals, might not be accurately matched.
One of the key findings of our research is that meticulous fine-tuning on one class of 3D shapes can lead to improved performance on other shape classes, reaching or surpassing the accuracy of supervised methods. We compare and discuss several fine-tuning strategies. Additionally, we delve deeply into how the scale of an object in a sketch influences the similarity of features at different network layers, helping us identify which network layers provide the most accurate matching.
Significantly, we discover that ViT and ResNet perform best when dealing with similar object scales. We believe that our work will have a significant impact on research in the sketch domain, providing insights and guidance on how to adopt large pretrained models as perceptual losses.
\end{abstract}

%%%%%%%%% BODY TEXT
\section{Introduction}
\label{sec:intro}
As image vision algorithms rapidly advance, we see a recent surge of interest in sketch understanding \cite{yang2021sketchgnn,qi2022one,li2022free2cad,chowdhury2023can,qu2023sketchxai,lin2023zero} and generation \cite{CLIPasso,chan2022learning}. Sketch is the earliest form of visual communication for humanity as a whole, as well as for each individual. 
However, for vision algorithms, it poses a number of challenges caused by the diversity of sketching styles, skills, and sketch sparsity. Sketches can be very abstract and visually different from photos. Each sketching scenario results in visually very different renditions. 
People can easily interpret sketches from very abstract to highly detailed or stylized, but is there an algorithm or model that can reliably handle all styles and scenarios?

Inspired by the success of so-called foundation models trained on large datasets in a range of zero-shot applications in the image domain \cite{michel2022text2mesh,saharia2022photorealistic,nichol2022glide, richardson2023texture,li2023clip},  several works exploit its application in the sketch domain.
There are a number of inspiring attempts of adapting CLIP (Contrastive Language-Image Pre-Training) \cite{clip} as a perceptual loss for deep model training \cite{frans2022clipdraw,schaldenbrand2021styleclipdraw,CLIPasso,vinker2022clipascene}, for model performance evaluation \cite{zheng2023locally}, or as a way to alleviate the need for the sketch data during training for a downstream task \cite{sanghi2023sketch}. However, are the used encoders able to discriminate fine-grained differences within a sketch domain or across sketch and image domains? Several works indicate that these models do not necessarily perform that well in a zero-shot sketch-to-image comparison \cite{fscoco,sain2023clip}. The works are then either residing to fine-tuning existing models \cite{fscoco,sain2023clip} or training from scratch by adding additional losses or sketch-targeted solutions \cite{lin2023zero,sangkloy2022sketch}. 

% Fine-tuning foundation models is an emerging field \cite{bahng2022visual,jia2022visual,zhou2022conditional}, and concurrent to our work by Sain et al.~\cite{sain2023clip} explores fine-tuning strategies for the tasks of sketch-based image retrieval. 
With our work, firstly, we aim to shed light on the ability of the popular pretrained models to discriminate individual 3D instances in their multi-modal 2D projections. To achieve this, we evaluate encoders trained with CLIP \cite{clip} and via a classification task training on the ImageNet dataset \cite{imagenet}. Namely, we study their performance in matching viewpoints and object identities in sketches and images. 
Secondly, we investigate alternative fine-tuning strategies, inspired by \cite{bahng2022visual,jia2022visual,zhou2022conditional}.
In our work, we compare visual prompt learning \cite{jia2022visual}, layer normalization weights learning \cite{frankle2020training,sain2023clip} with a careful fine-tuning of all weights. 
Thirdly, we show that well-designed fine-tuning on a single shape class can lead to improved performance on other shape classes, sometimes
surpassing the accuracy of supervised methods. Importantly, we show that fine-tuning can be done on synthetically-generated sketches for a set of 3D shapes without the requirement to use freehand sketches. We demonstrate the generalization of our approach to relatively abstract freehand sketches from the AmateurSketch dataset \cite{qi2021toward}. 
We refer to this scenario as \emph{fine-tuned but zero-shot}. 
As a test application, we consider sketch-based 3D shape retrieval. Effectively, we introduce the first sketch-based 3D shape retrieval method with state-of-the-art performance that does not require per-class training or fine-tuning. Our fine-tuning only requires a set of 3D shapes of just one category. It is a reasonable assumption for the 3D shape retrieval task.

Fourthly, we perform a detailed performance analysis of different layers of ViT and ResNet-based encoders pretrained either with CLIP training or with a classification task on the ImageNet dataset. We study how the line width and object scale affect performance and find that similar settings can be considered optimal for ViT and ResNet-based encoders. We note that most works \cite{fscoco,sain2023clip,lin2023zero}  use the activation of the final layer of an encoder. Yael et al. \cite{CLIPasso} observed that while these features excel at capturing semantic meaning, intermediate layers are more suitable when comparing spatial structures. In our research, we offer an in-depth analysis with regard to our specific problem.

In summary, our key contributions are:
\begin{itemize}
    \item A comprehensive study of the ability of the popular pretrained encoders to discriminate individual 3D instances in their multi-modal 2D projections;
    \item Extensive analysis of the similarity estimation performance using various layer features and exploration of the impact of the object's scale;
    \item Comparison of various fine-tuning strategies on the task of matching sketches in distinctive sketch styles;
    \item Fine-tuning approach that requires as little as a set of 3D shapes of a single category, and generalizes to freehand sketches and other shape class categories, reaching the performance of state-of-the-art fully supervised methods. 
\end{itemize}

\begin{figure*}[t]
\vspace{-0.2cm}
  \centering
    \includegraphics[width=\linewidth]{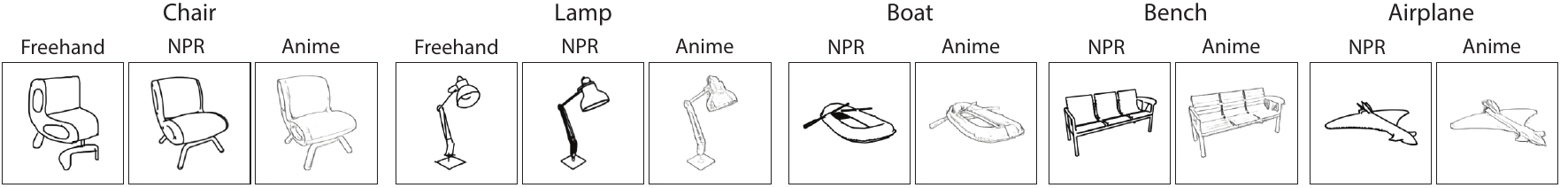}
   \caption{Examples of sketches in our datasets. Please see \cref{sec:data} for details.}
   \label{fig:styles}
\end{figure*}

\section{Related work}
\label{sec:rel_work}

\subsection{Sketch-based 3D shape retrieval}
%\paragraph{Shape retrieval} 

\subsubsection{Category-level and fine-grained retrieval}
Most of the works in sketch-based 3D model retrieval \cite{eitz2012sketch,yoon2010sketch,lu2013efficient,li2014shrec,wang2015sketch,  zhao2015learning, li2016model, zhu2016learning, xie2017learning,  dai2017deep,he2018triplet,qi2018BMVC, lei2019deep, chen2019deep, xu2020sketch, chen2022self, zhao2022jfln, xu2022domain} focus on the problem of \emph{category level} retrieval: They aim to retrieve any instance of a particular object category. 
In other words, the retrieval is considered to be successful if, given a sketch of an object, the retrieved top $N$ 3D models belong to the same category. 
%Chen et al.~\cite{chen2022self} build on the assumption that views of the same object should be projected closer to each other in the feature space than views of other models.

Only two works \cite{qi2021toward,chen2023spatially} addressed fine-grained sketch-based 3D model retrieval in a supervised setting.
Qi et al.~\cite{qi2021toward} collected the first dataset of instance-level paired freehand sketches and 3D models, which we also use to test our model.
They use triplet loss training \cite{wang2014learning}, classic for retrieval tasks, and represent 3D models using multi-view RGB renderings. 
The main novelty of their paper lies in learning view attention vectors.
In concurrent to our work, Chen et al.~\cite{chen2023spatially} train and test on the data by Qi et al.~\cite{qi2021toward}. 
Unlike \cite{qi2021toward}, they learn to project all sketch views to the same latent representation. The main performance gain is caused by dividing the images into three parts and learning to match the features of each part individually. Unlike both of these works, we represent 3D shapes using NPR renderings rather than RGB images. 

% and \emph{do not use freehand} sketches for our zero-shot or fine-tuned models. 

To reduce the domain gap between sketch queries and 3D models, several works \cite{luo2021fine,luo2022structure} study fine-grained retrieval from a 3D sketch created while wearing a virtual reality headset.
Our model aims for much more accessible inputs that can be created with a computer mouse or on paper.

\subsubsection{Multi-view feature aggregation}
%%% Features aggregation strategies:
% \paragraph{Multi-view feature aggregation} 
Like majority of the works on sketch-based 3D model retrieval, we use multi-view shape representation, however many of these works differ in how they aggregate features across viewpoints.
Thus, Xie et al.~\cite{xie2017learning} use the Wasserstein barycentric of 3D shapes projections in the CNN feature space to represent 3D shapes.
He et al.~\cite{he2018triplet} follow MVCNN~\cite{su2015multi} and aggregate views features with element-wise maximum operation across the views in the view-pooling layer. 
Lei et al.~\cite{lei2019deep} proposed a representative view selection module that aims to merge redundant features for similar views.
Chen et al.~\cite{chen2019deep} learn multi-view feature scaling vectors which are applied prior to average pooling vector, in order to deal with non-aligned 3D shape collections. 
Qi et al.~\cite{qi2021toward} learn view attention vectors conditioned on the input sketch, which allow to reduce the domain gap between a sketch and multi-view projections of a 3D shape. 
Zhao et al.~\cite{zhao2022jfln} leverages spatial attention \cite{zhang2018self} to exploit view correlations for more discriminative shape representation. 
%
% We explore here a simpler strategy that we describe in \cref{sec:method_views}.
In our work, we focus on learning view features that can be used to find the correct shape identity and view across different sketch styles: e.g. freehand and NPR. 

\subsection{Multi-modal retrieval}
Multi-modal retrieval is not directly related to our work, but two concurrent works \cite{schlachter2022zero,sangkloy2022sketch} are worth mentioning as they rely on encoders pretrained with the CLIP model. 
They explore CLIP embeddings for retrieval from multi-modal inputs such as 2D sketches or images and text. 
Sangkloy et al.~\cite{sangkloy2022sketch} study image retrieval and focus on fine-tuning CLIP using triplets of synthetic sketches, images, and their captions. 
They rely on the availability of textual descriptions matching their images, while we require only the availability of 3D shapes from just one 3D shape class.
Similarly to us, Schlachte et al.~\cite{schlachter2022zero} study zero-shot 3D model retrieval using the CLIP model, but only explore the weighted fusion of CLIP features from multiple inputs for artistic control. 
Unlike them, we perform an in-depth study analyzing object scale, feature layers, and fine-tuning strategies. 

% \todo{In our work, we conduct a comprehensive study of viewpoints selection and in the supplementary we show that this is not an optimal viewpoint selection strategy.} 
% Instead, in our work, we focus on determining the optimal strategy for 3D shape retrieval from freehand sketch inputs using CLIP.
% \todo{Extend describing recent work}

\subsection{Sketch datasets}
With the advent of sketch datasets \cite{eitz2012sketch,lu2013efficient,antol2014zero,sangkloy2016sketchy,ha2018neural,gryaditskaya2019opensketch,qi2021toward,fscoco} the research on sketching thrives.
However, it is costly and challenging to collect a dataset of freehand sketches, especially when there is a requirement for instance-level pairing between several domains. 
The common practice is to let participants study a reference image for a short period of time and then let them draw from memory \cite{eitz2012sketch,antol2014zero,sangkloy2016sketchy,fscoco,sangkloy2022sketch}. 
This task becomes increasingly challenging when the pairing is required to be between 3D shapes and sketches, as one has to ensure that the viewpoints are representative of those that people are more likely to sketch from \cite{lu2013efficient, zhao2015learning,xu2020sketch,qi2021toward,gryaditskaya2019opensketch}.

To the best of our knowledge, there is only one dataset \cite{qi2021toward} of freehand sketches by participants with no prior art experience paired with 3D shapes, that takes views into account and follows the protocol of sketching from memory.  
The small dataset collected by Zhang et al.~\cite{zhang2021sketch2model} for each object contains only one sketch viewpoint, and the viewpoints are non-representative, they are uniformly sampled around 3D shapes. It contains too few examples and is too noisy for retrieval performance evaluation. 
The recent dataset of paired sketches and 3D models of cars \cite{guillard2021sketch2mesh} similarly was collected without taking into account viewpoints preferences, and  the sketches are drawn directly on top of image views and mostly contain outer shape contours. 
We, therefore, evaluate our approach on the dataset by Qi et al.~\cite{qi2021toward}, as the only existing representative dataset with instance-level pairing between sketches and 3D shapes.

%\paragraph{Image retrieval}

% \subsection{Perceptual losses}
\section{Method}
\label{sec:method}

In this and the following sections, we present our method for zero-shot sketch-based 3D retrieval. We then provide a comparison to alternative strategies in \cref{ablation}. 
To enable sketch-based 3D shape retrieval, we represent 3D shapes using their multi-view projections, commonly used in sketch-based retrieval \cite{xie2017learning,he2018triplet,lei2019deep,chen2019deep,zhao2022jfln,lu2013efficient, zhao2015learning,xu2020sketch}. 
To reduce the domain gap, we use NPR renderings instead of RGB renderings for multi-view 3D shape representation.
In the supplemental, we provide a detailed study of the ability of the popular pretrained models to discriminate individual 3D instances in their multi-modal 2D projections: we compare RGB renderings, NPR renderings, and freehand sketches.
% \cite{qi2021toward}.
% Following previous literature, we rely on the assumption that for each class of 3D shapes, there are several viewpoints from which humans are more likely to depict an object \cite{lu2013efficient, zhao2015learning,xu2020sketch,qi2021toward,gryaditskaya2019opensketch}.

% We represent 3D shapes with their multi-view projections, using NPR rendering (we provide implementation details in \cref{sec:shape-details}).

\subsection{Zero-shot}
\label{sec:method_zero_shot}
Given an encoder, trained on a pretext task, we first compute embeddings of a Query sketch $Q$ and Gallery 3D shape $G$ views using features of a chosen encoder's layer. 
We then assign the similarity between a sketch and a 3D shape as the maximum cosine similarity between a sketch embedding and individual 3D shape views embeddings.
Formally, this can be written as follows:
\begin{equation}
    \mathrm{sim}(Q,G) = \max_{v \in views}\mathrm{d}(\mathrm{E_\ell}(Q), \mathrm{E_\ell}(G_v)),
    \label{eq:zero_shot}
\end{equation}
$G_v$ is a 3D shape view, $E_l(\cdot)$ denotes layer $\ell$ features extracted with the encoder $E$ and $d$ stands for the cosine similarity\footnote{We experimented with the Mean Squared Error (MSE) distance, taking the minimum MSE distance between a query and shape individual views. We have not observed an obvious advantage of one other another.}. 

We center and scale 3D objects in query and shape views to fit the same bounding box in both representations.

\subsection{Fine-tuned but zero-shot }
\label{sec:method_zero_shot_but_fine_tunded}

We propose a contrastive view-based fine-tuning approach that leverages synthetically-generated sketches of single or multiple 3D shape classes.
We represent all available 3D shapes with $V$ views, using two different approaches to synthetic sketch generation: model-based \cite{freestyle} and view-based \cite{chan2022learning} NPR algorithms. 
We then adapt CLIP contrastive loss \cite{clip,sohn2016improved} to match identical shape views in these two synthetic sketch styles. 

Namely, given a batch with $B$ objects, we randomly select one view in two styles for each.
We then compute the pairwise weighted dot product between any two views in two different styles:
\begin{equation}
     s_{i,j} := s(G_i^{st1},G_j^{st2}) :=  e^t <\mathrm{E_\ell}(G_i^{st1}), \mathrm{E_\ell}(G_j^{st2})>,
    \label{eq:dot_product}
\end{equation}
$<\cdot,\cdot>$ is a dot product, $G_i^{st}$ is some view of the $i$-th object in the mini-batch in one of two styles, and $t$ is a learned parameter. We provide details on styles and views in \cref{sec:shape-details}.

We then compute the following contrastive loss:

\begin{align}
    \mathcal{L} = -\frac{1}{2B}\sum_{i=1}^B\left( \log \frac{\exp(s_{i,i})}{\sum_{j=1}^B \exp(s_{i,j})  } + \right. \nonumber \\
   \left.
    \log \frac{\exp(s_{i,i})}{\sum_{j=1}^B \exp(s_{j,i})  }
    \right).
    \label{eq:contrastive_loss}
\end{align}
%
% Namely, given a batch with $B$ objects, we randomly select one view in two styles for each. We compute the pairwise weighted dot product between any two views in two different styles:
% \begin{equation}
%      s(I_{st1},J_{st2}) =  e^t <\mathrm{E_\ell}(I_{st1}), \mathrm{E_\ell}(J_{st2})>,
%     \label{eq:dot_product}
% \end{equation}
% $<\cdot,\cdot>$ is a dot product, $I_{st1}$ and $J_{st2}$ are any two views in two different styles, and $t$ is a learned parameter.
%
%
% We then compute a symmetric cross-entropy loss:
% \begin{align}
%     % \mathcal{L}_{st1} & = cross\_entropy(S, labels) \\
%     \mathcal{L}_{st1} & = -\frac{1}{B}\sum_{o=1}^B \sum_{c=1}^By_{o,c}\log(\frac{e^{S_{o,c}}}{\sum_{j=1}^B e^{S_{o,j}}}) \\
%     % \mathcal{L}_{st2} & = cross\_entropy(S^\intercal, labels) \\
%     \mathcal{L}_{st1} & = -\frac{1}{B}\sum_{o=1}^B \sum_{c=1}^By_{o,c}\log(\frac{e^{S^\intercal_{o,c}}}{\sum_{j=1}^B e^{S^\intercal_{o,j}}}) \\
%     % \mathcal{L}_{st1} & = -\sum_{e=1}^B \sum_{c=1}^By_{e,c}\log(S^\intercal_{e,c}) \\
%     \mathcal{L} & =  (\mathcal{L}_{st1} + \mathcal{L}_{st2}) / 2
%     \label{eq:contrastive_loss}
% \end{align}
% where $B$ is the batch size, $y$ is a matrix of one-hot encodings from class $0$ to class $B$, indicating shape identity, and $S$ is a matrix of pairwise view similarities.  
% where $labels$ is a vector containing the integer values between $0$ and $B$, indicating shape identity, and $S$ is a matrix of pairwise view similarities.  
%
% optimized to identify a positive example from multiple
% negative examples. 
Due to our batch construction, this objective trains the network to produce features such that the same views of the same object in different styles have similar embeddings. 
This objective neither pushes different views of the same object to have identical embeddings nor pushes them apart. Fine-tuning updates the weights of the visual encoder and the temperature parameter $t$.

Note that \cref{eq:zero_shot,eq:dot_product} can be computed based on the features from any layer and not only the final one. 
In this case, we only updated the weights up to the layer whose features we use to compute similarity.

\begin{figure*}[!t]
  \centering
    \includegraphics[width=\linewidth]{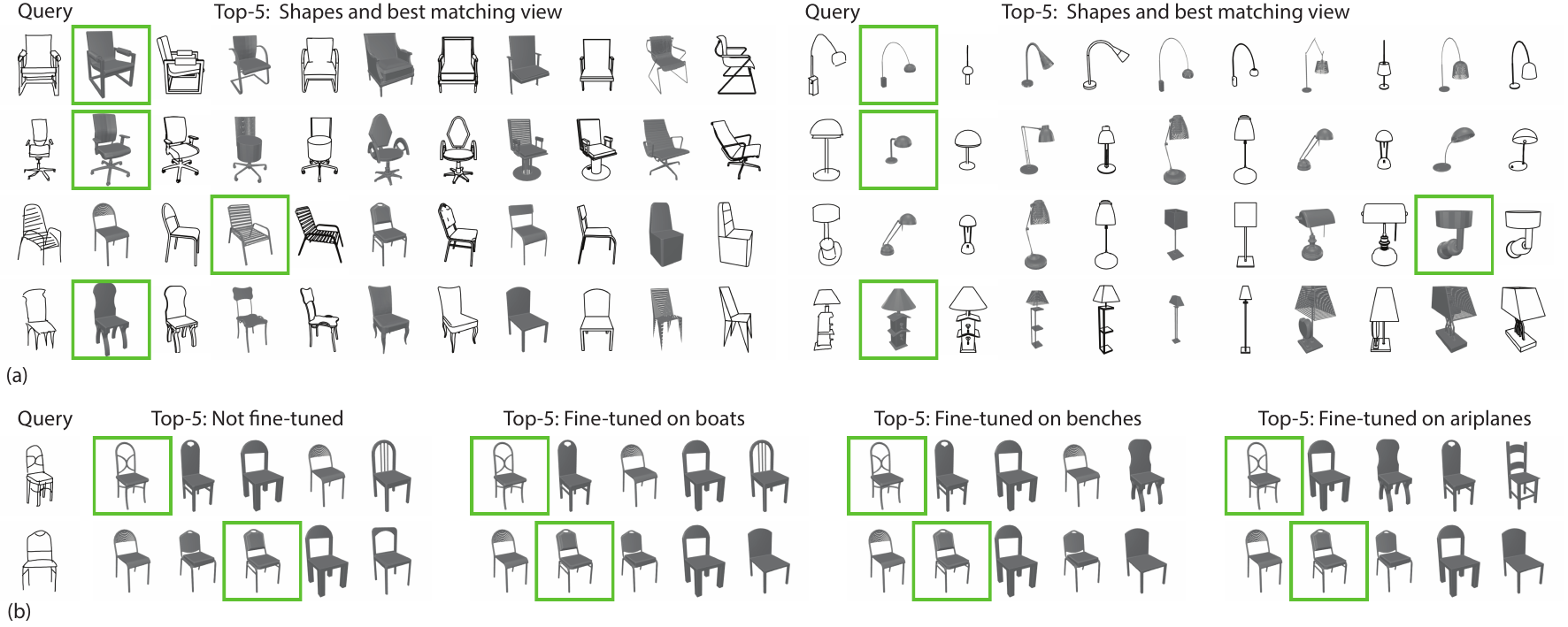}
    \vspace{-0.5cm}
   \caption{Qualitative results obtained with features of the 6th layer of the ViT encoder pretrained on CLIP and fine-tuned using our method. The queries are freehand sketches from the AmateurSketch dataset \cite{qi2021toward}. Green boxes highlight groundtruth shapes. (a) shows retrieved shapes and the best matching view according to \cref{eq:zero_shot}; (b) shows retrieval results without our fine-tuning and with fine-tuning on each of the free classes: boats, benches, and airplanes.}
   \label{fig:results}
   \vspace{-0.3cm}
\end{figure*}

\section{Implementation details}
\label{sec:implementation_details}

\subsection{Encoder}
In the default setting, as an encoder, we use ViT pretrained with CLIP. We compute similarity using the 6-th layer.

\subsection{Datasets} 
\label{sec:data}
We use two types of datasets: (1) the dataset of freehand sketches by participants without art experience, and (2) the dataset of synthetically generated sketches in two styles for 11 classes of the ShapeNet 3D shape dataset \cite{shapenet}. Different styles are shown in \cref{fig:styles}, and described in detail below.

\subsubsection{Freehand sketches} 
We use the dataset of freehand sketches by Qi et al.~\cite{qi2021toward} to evaluate the models' performance.
This dataset contains sketches for two shape categories: \chair{} and \lamp{}, representing 1,005 and 555 3D shapes from the respective class of the ShapeNet dataset \cite{shapenet}. 
The sketches are created by participants without any prior sketching experience, and fit well the scenario we are targeting. 
The sketches are drawn from a viewpoint with a zenith angle of around 20 degrees. 
For each category three settings of azimuth angles are used. 
For the \chair{} category, they are $0^\circ$, $30^\circ$ and $75^\circ$, while for the \lamp{} category they are $0^\circ$, $45^\circ$ and $90^\circ$.
These particular viewpoints are selected based on sketching literature and pilot studies, as the most likely viewpoints.

The dataset provides a split to training, validation, and test data. 
To facilitate comparison with previous supervised work, we only use a test set of sketches to test models. The test set consists of 201 and 111 sketch-3D shape quadruplets for the \chair{} and \lamp{} categories, respectively.
We do not use any freehand sketches for training. 
Prior to testing, we re-scale and center objects' projections in freehand sketches to occupy the central image area of $129\times129$.

\subsubsection{Synthetic sketches} 
\label{sec:synhteic_styles}
Additionally, we create a dataset of synthetic sketches in two styles, representing 3D shapes from the ShapeNetCore 3D shape dataset \cite{shapenet}. 
We select 11 of the 13 ShapeNetCore classes, discarding two classes with the lowest number of 3D shapes.

\paragraph*{Views and camera setting}
We follow camera settings used to collect sketches in the dataset of freehand sketches \cite{qi2021toward}. In particular, we use for all shape classes viewpoints with the following azimuth angles: $0^\circ, 30^\circ, 45^\circ, 75^\circ$ and $90^\circ$. We set the camera distance to an object to $2.5$ and the camera zenith angle to $20^\circ$. 
The size of rendered views is $224 \times 224$ unless specified otherwise.

\paragraph*{NPR (style-1)} 
We render views using silhouettes and creases lines in Blender Freestyle \cite{freestyle}. 
We render views as SVGs and then re-scale and center objects' projections in freehand sketches to occupy the central image area of $129\times129$. Prior to rasterization, we assign each stroke a uniform stroke width of $2.2px$.

\paragraph*{Anime (style-2)} 
We obtain the second synthetic sketch style by first rendering RGB images of 3D shapes using Blender Freestyle with the same camera settings as for the first NPR synthetic style. 
We then re-scale and center objects' projections in RGB renderings to occupy the central image area of $129\times129$.
Finally, we generate synthetic sketches in the second style, using the pre-trained network \cite{chan2022learning} in \emph{anime} style.

\subsection{Data usage}
\label{sec:data-usage}

\subsubsection{3D Shape representation}
\label{sec:shape-details}
We represent a 3D shape with its multi-view NPR projections: We use the set of $0^\circ,30^\circ,45^\circ,75^\circ$ and $90^\circ$, common for chair and lamp category sketches from the \emph{AmateurSketch} dataset \cite{qi2021toward}, to represent 3D models.

\subsubsection{Fine-tuning}
\label{sec:fine-tuning-details}
We split 3D shapes in each class into training (70 \%), validation (15\%), and test (15\%) sets.
% When training on multiple classes, we unite their training and validation sets.
We set the learning rate to $10^{-7}$, the batch size to 64, and use the Adam optimizer. 
\emph{We note that the choice of the learning rate is critical, as larger learning rates will result in overfitting harming the performance.}

\paragraph*{Data augmentation}
While fine-tuning, we augment synthetic sketches in the anime style with random affine transformation, translation, rotation, and scaling operations. 
This augmentation simulates the type of distortions that we can encounter in freehand query sketches. Even if we scale and center objects in freehand sketches in processing, sketches might contain small rotations.
The translation moves an image along the $x$ and $y$ axes for a random number of pixels in the range $[-10\%, +10\%]$ of the image size. 
The rotation is sampled between $[-10, +10]$ degrees.
Finally, we increase or decrease the object's bounding box size by a random value in the range [-10\%, +10\%] of the image size.

\paragraph*{Checkpoint selection}
We train our fine-tuning model for 500 epochs.
At test time we use the weights from the last epoch. 

\subsubsection{Test time}
We test our retrieval models on the freehand sketches. 
We also test on synthetic sketches to show generalization to other shape classes.
By default, we use sketches in the anime style with azimuth angles set to $0^\circ$, $45^\circ$, and $90^\circ$ as queries. 
To facilitate comparisons with performance on freehand sketches, for each shape class we form the final test sets by randomly selecting just 200 3D shapes non-overlapping with training or validation sets.

\begin{table}[t]
\centering
\resizebox{\columnwidth}{!}{%
\begin{tabular}{l|rr|rr|cc|}
 & \multicolumn{2}{c|}{Chairs} & \multicolumn{2}{c|}{Lamps} & \multicolumn{2}{c|}{\begin{tabular}[c]{@{}c@{}}Avg. score. \\ Anime $\rightarrow$ NPR\end{tabular}} \\ \hline
Method & \multicolumn{1}{c}{acc@1} & \multicolumn{1}{c|}{acc@5} & \multicolumn{1}{c}{acc@1} & \multicolumn{1}{c|}{acc@5} & acc@1 & acc@5 \\ \hline
\cite{qi2021toward} & 56.72 & 87.06 & 57.66 & 87.39 & n.a. & n.a. \\
\cite{chen2023spatially} & \textbf{83.08} & \textbf{97.01} & {\ul 78.08} & \textbf{95.50} & n.a. & n.a. \\
\hline
ViT-CLIP L-6 & 74.79 & 89.39 & 73.27 & 89.49 & \multicolumn{1}{r}{82.48} & \multicolumn{1}{r|}{93.82} \\
ViT-CLIP* L-6 & {\ul 77.11} & {\ul 92.32} & \textbf{78.38} & {\ul 92.39} & \multicolumn{1}{r}{\textbf{87.84}} & \multicolumn{1}{r|}{\textbf{97.13}} \\
\end{tabular}%
}
\vspace{0.05cm} 
\caption{Our zero-shot results versus supervised methods: \cite{qi2021toward} and concurrent to our work \cite{chen2023spatially}. Neither \cite{qi2021toward} nor \cite{chen2023spatially} provide code, therefore, we use the numbers provided in their respective papers.
For the \emph{ViT-CLIP} methods, we center and scale objects in reference and query views according to optimal scaling. 
\emph{L-6} indicates the layer whose features we use for similarity computation.  
\emph{ViT-CLIP*} represent the average performance results of three individual fine-tuning experiments on the three classes: \emph{boat}, \emph{airplane}, and \emph{bench}, using synthetic sketches.  \emph{Avg. score. anime} represents average results on 11 classes where queries are in anime style and gallery shapes are represented using muti-view NPR projections. The boldface font highlights the best results, and the underscore highlights the second-best results.
}
\label{tab:comparison}
\end{table}

\begin{figure*}[t]
\includegraphics[width=\linewidth]{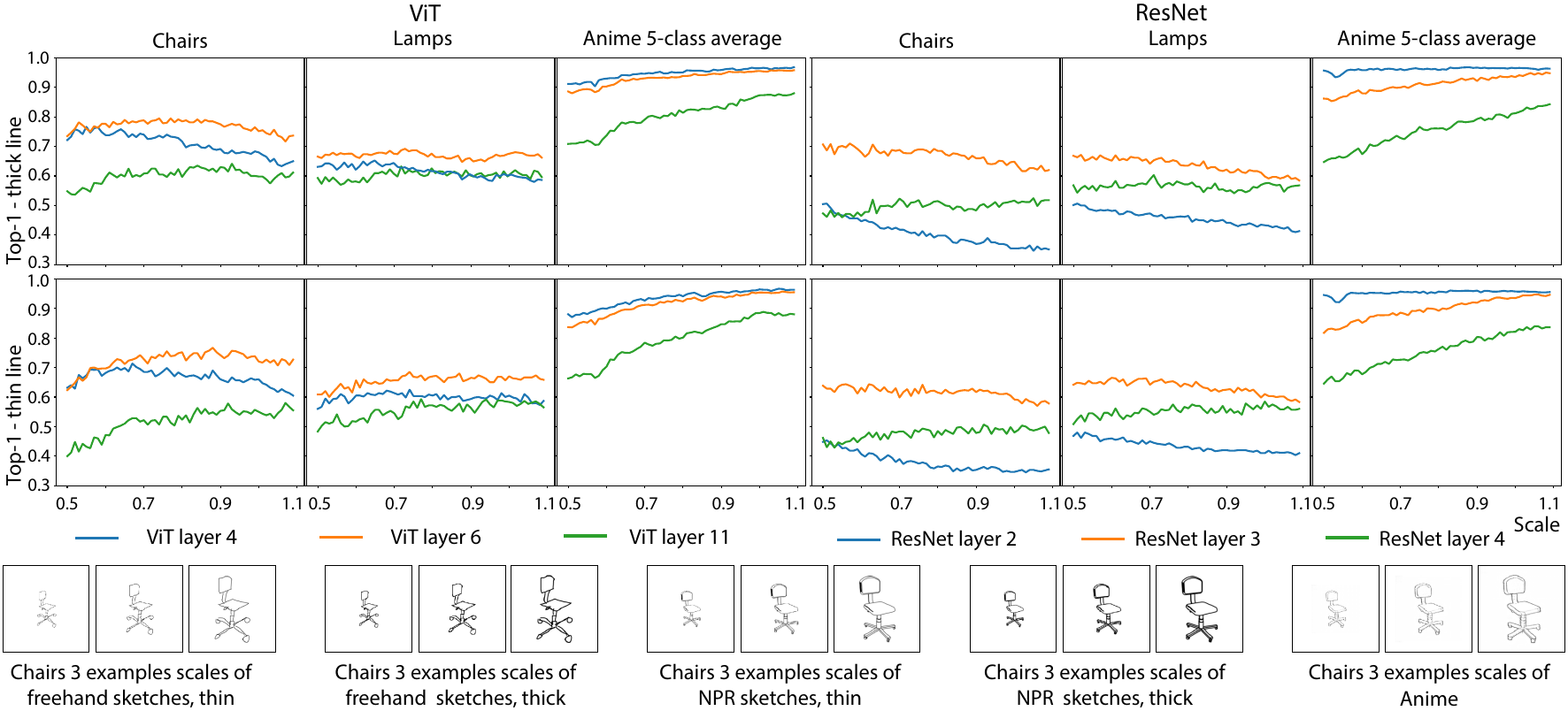}
\vspace{-0.5cm}
\caption{Role of the object projection area, line width, and feature layer in the ability to predict similarity between views in different domains. Please see \cref{sec:bounding_box} for the details.}
\label{fig:OBB_layer}
\end{figure*}
\section{Results}
To evaluate retrieval accuracy, we use the standard for retrieval tasks Top-1 (Acc@1), and Top-5 (Acc@5) accuracy measures. 
They evaluate the percentage of times the ground-truth is returned among the top 1 and top 5 ranked retrieval results, respectively.  

\cref{tab:comparison} (\emph{ViT-CLIP L-6}) shows the retrieval accuracy of our zero-shot setting on the freehand sketches and synthetic sketches in anime style. 
We then perform three individual fine-tuning experiments on three classes: \emph{boat}, \emph{airplane}, and \emph{bench}, using synthetic sketches, and report an average accuracy over the three experiments in \cref{tab:comparison} (\emph{ViT-CLIP* L-6}).
We compare with two supervised works by Qi et al. \cite{qi2021toward} and Chen et al.\cite{chen2023spatially} who train on one class at a time and use freehand sketches from \cite{qi2021toward}.
As no code is available for the competitors, we report the numbers from their respective papers. 

Our zero-shot models are able to achieve remarkable results, surpassing \cite{qi2021toward} in all respects. 
This shows the generalization ability of our method to different styles and diverse shape classes. 
Compared to concurrent to our work \cite{chen2023spatially}, we can see that accuracy of our zero-shot method can be further improved. 
Note that on the \lamp{} category we outperform the concurrent supervised method in top-1 accuracy, while our method is zero-shot!
Our \emph{fine-tuning but zero-shot} improves top-1/5 retrieval accuracy on average by $4.3$ and $3$ points, respectively, over zero-shot performance.
The visual results for our method are shown in \cref{fig:results}.

\section{Ablation studies}
\label{ablation}
\subsection{Choice of an encoder and pretext task}
\label{sec:choice}
In this section, we compare (1) two types of encoders: ViT and ResNet, and (2) two types of pretext tasks: CLIP training and classification task training on the ImageNet dataset \cite{imagenet}.
All the models share the same input image size of $224\times224$ except for ViT pre-trained on ImageNet. In the latter case, the image size is $384\times384$, and we re-scale and center objects’ projections in freehand and synthetic sketches to occupy the central area of $291\times291$.

\cref{tab:choice} shows the comparison in retrieval accuracy in a zero-shot setting without fine-tuning for several best-performing layers. 
It shows that ViT encoder pretrained with CLIP model achieves the best results, and justifies the use of it as our default in most of the experiments.  
Interstingly, training on the ImageNet for the RestNet encoder gives slightly better performance than training with the CLIP model.

\begin{table}[]
\centering
\resizebox{\columnwidth}{!}{%
\begin{tabular}{l|rr|rr|rr|}
 & \multicolumn{2}{c|}{Chairs $\rightarrow$ NPR} & \multicolumn{2}{c|}{Lamps $\rightarrow$ NPR} & \multicolumn{2}{c|}{\begin{tabular}[c]{@{}c@{}}Avg.score \\ Anime $\rightarrow$ NPR\end{tabular}} \\ \hline
Method & \multicolumn{1}{c}{acc@1} & \multicolumn{1}{c|}{\cellcolor[HTML]{EFEFEF}acc@5} & \multicolumn{1}{c}{acc@1} & \multicolumn{1}{c|}{\cellcolor[HTML]{EFEFEF}acc@5} & \multicolumn{1}{c}{acc@1} & \multicolumn{1}{c|}{\cellcolor[HTML]{EFEFEF}acc@5} \\ \hline
ViT CLIP L-6 & \textbf{74.79} & \cellcolor[HTML]{EFEFEF}\textbf{89.39} & \textbf{73.27} & \cellcolor[HTML]{EFEFEF}\textbf{89.49} & {\ul{82.48}} & \cellcolor[HTML]{EFEFEF}{93.82} \\
ViT ImageNet L-5 & 63.35 & \cellcolor[HTML]{EFEFEF}82.92 & 66.67 & \cellcolor[HTML]{EFEFEF}87.99 & 81.77 & \cellcolor[HTML]{EFEFEF}{\ul 94.14} \\ \hline
ResNet CLIP L-3 & 65.17 & \cellcolor[HTML]{EFEFEF}{\ul 84.08} & {\ul 69.97} & \cellcolor[HTML]{EFEFEF}87.99 & 76.97 & \cellcolor[HTML]{EFEFEF}90.36 \\
ResNet ImageNet L-3 & {\ul 66.50} & \cellcolor[HTML]{EFEFEF}82.09 & 65.77 & \cellcolor[HTML]{EFEFEF}{\ul 88.89} & {\textbf{83.82}} & \cellcolor[HTML]{EFEFEF}{\textbf{95.29}}
\end{tabular}%
}
\vspace{0.05cm} \caption{Comparison of ResNet and ViT encoders trained either with CLIP model or classification task on the ImageNet dataset. See \cref{sec:choice} for the details. In all cases, objects in sketches are optimally scaled and centered.}
\label{tab:choice}
\vspace{-0.5cm}
\end{table}

\subsection{Object projection area, line width and feature layer}
\label{sec:bounding_box}
In our preliminary experiments, we observed that scaling sketches and 3D model projections to fit the same bounding box area results in improved retrieval accuracy (See \cref{tab:alignment}). These findings also align with the experiments in \cite{chen2023spatially}. 
We then are interested in how sensitive different backbones (ViT and ResNet) are to (1) the scale of the object in the image plane; (2) the line width, and  (3) how the accuracy of feature similarities according to features from different layers varies with object scale.

\begin{table}[ht]
\centering 
\resizebox{1.0\columnwidth}{!}{
\scriptsize{
\begin{tabular}{l|crrr}
        & \multicolumn{2}{c|}{Chair}                                               & \multicolumn{2}{c}{Lamp}                              \\
        & \multicolumn{1}{l}{acc@1}          & \multicolumn{1}{l|}{acc@5}          & \multicolumn{1}{l}{acc@1} & \multicolumn{1}{l}{acc@5} \\ \hline
ViT-CLIP L-6 w/o alignment & \multicolumn{1}{r}{{69.82}} & \multicolumn{1}{r|}{{86.40}} & {67.27}            & {87.69}            \\ \hline
ViT-CLIP L-6 & \multicolumn{1}{r}{\textbf{74.93}} & \multicolumn{1}{r|}{\textbf{89.39}} & \textbf{73.27}            & \textbf{89.49}            \\\hline
\end{tabular}%
}}
\vspace{0.05cm} \caption{Comparison of the zero-shot retrieval performance of the ViT encoder trained with CLIP on the datasets without objects centering and rescaling vs.~on the datasets where objects in sketches are centered and scaled as described in \cref{sec:data-usage}.
} 
\label{tab:alignment}
\vspace{0pt}
\end{table}

\begin{table*}[th!]
\vspace{-0.4cm}
\centering
 \resizebox{\textwidth}{!}{%
\begin{tabular}{l|cc|cc||cc|cc||cc|cc|}
 & \multicolumn{2}{c|}{Chairs $\rightarrow$ NPR} & \multicolumn{2}{c||}{Chairs $\rightarrow$ Anime} & \multicolumn{2}{c|}{Lamps $\rightarrow$ NPR} & \multicolumn{2}{c||}{Lamps $\rightarrow$ Anime} & \multicolumn{2}{c|}{\begin{tabular}[c]{@{}c@{}}Avg.score \\ Anime $\rightarrow$ NPR\end{tabular}} & \multicolumn{2}{c|}{\begin{tabular}[c]{@{}c@{}}Avg.score \\ NPR $\rightarrow$ Anime\end{tabular}} \\ \hline
Method & acc@1 & \cellcolor[HTML]{EFEFEF}acc@5 & acc@1 & \cellcolor[HTML]{EFEFEF}acc@5 & acc@1 & \cellcolor[HTML]{EFEFEF}acc@5 & acc@1 & \cellcolor[HTML]{EFEFEF}acc@5 & acc@1 & \cellcolor[HTML]{EFEFEF}acc@5 & acc@1 & \cellcolor[HTML]{EFEFEF}acc@5 \\ \hline
ViT CLIP L-6 & \multicolumn{1}{r}{\textbf{74.79}} & \multicolumn{1}{r|}{\cellcolor[HTML]{EFEFEF}\textbf{89.39}} & \multicolumn{1}{r}{63.35} & \multicolumn{1}{r||}{\cellcolor[HTML]{EFEFEF}80.93} & \multicolumn{1}{r}{\textbf{73.27}} & \multicolumn{1}{r|}{\cellcolor[HTML]{EFEFEF}\textbf{89.49}} & \multicolumn{1}{r}{62.16} & \multicolumn{1}{r||}{\cellcolor[HTML]{EFEFEF}82.88} & \multicolumn{1}{r}{\textbf{82.48}} & \multicolumn{1}{r|}{\cellcolor[HTML]{EFEFEF}\textbf{93.82}} & \multicolumn{1}{r}{77.03} & \multicolumn{1}{r|}{\cellcolor[HTML]{EFEFEF}90.94}
\end{tabular}%
 }
 \vspace{0.05cm} \caption{NPR vs. Anime 3D shape representation. In the notation $X \rightarrow Y$, $X$ is a query domain and $Y$ is a 3D shape representation domain.} %\todo{Fill in correct numbers}}
\label{tab:NPR_or_anime}
\end{table*}

\subsubsection{Object bounding box size \& line width}
We first obtain an initial common bounding box size ($170\times170$) by taking the smallest square bounding box that fully encompasses objects in all sketches in the dataset of freehand sketches in the form they are provided by Qi et al.~\cite{qi2022one}.
We rescale and center all object projections in all freehand and synthetic SVG sketch versions to this bounding box.  
We then use two settings of line width: thick (set to 2.2px) and thin (set to 1.0px) that we assign to all strokes (\cref{fig:OBB_layer}: 1st vs.~2nd rows), and rasterize the sketches.
We evaluate varying scaling of the original $170\times170$ bounding box size, by rescaling raster images so that the object projections are within varying bounding box sizes from $85\times85$ to $187\times187$ with 60 uniform steps (\cref{fig:OBB_layer}: scale in horizontal axes).

First, we observe that among the two considered line settings, thicker lines usage results in better retrieval accuracy. 
For freehand sketches, scaling between 0.7 and 0.8 that represents bounding boxes with sizes $119\times119$ and 
$136\times136$,
% $221\times221$, 
respectively, result on average in top performance across encoder architectures and feature layers. 
For synthetic sketches, the large the object in a sketch is, the more accurate is the prediction. We believe that is caused by two factors (1) the great degree of spatial alignment between two types of synthetic sketches and (2) the presence of very thin lines in anime style sketches at smaller object scales.

\subsubsection{Feature layers}
We study how retrieval accuracy varies when feature similarity is computed on features from different layers for different object projections bounding box sizes.  
In \cref{fig:OBB_layer}, we plot accuracy for similarity in \cref{eq:zero_shot} computed with features from layers $4$, $6$ and $11$ of the ViT encoder, and layers $2$, $3$ and 4 of the ResNet, both trained with CLIP (\cref{fig:OBB_layer}: first three columns vs.~last three columns).

\cref{fig:OBB_layer} shows that on the two categories of the dataset of freehand sketches features from mid-layers -- ViT layer 6 and ResNet layer 3 -- result in the best performance for both architectures.
On synthetic sketches, slightly better performance is achieved with features from lower layers: layer 4 of ViT and layer 2 of ResNet.
It can be also observed that for lower layers (ViT layer 4 and ResNet layer 2) the performance is increasing as object area is decreasing, while for higher layers (ViT layer 11 and ResNet layer 4) the behavior is opposite. 
The intuition is that the features from higher layers are better suited for more abstract sketches and extracting sketch semantic meaning, while lower layers focus more on spatial details. 
Indeed, as NPR and anime synthetic sketches are spatially more similar than NPR and freehand sketches, the lower layers result in better performance when anime sketch is used as a query. 

\begin{figure}[t]
\centering
    \includegraphics[width=\columnwidth]{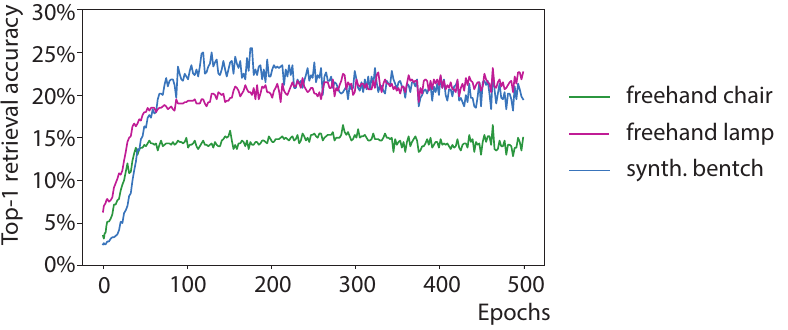}
\vspace{-0.7cm}
\vspace{0.1cm} \caption{Top-1 retrieval accuracy vs.~epoch number, when ViT encoder is trained from scratch as described in \cref{sec:scratch} on synthetic sketches of the \bench{} class.}
\label{fig:from_scratch}
% \vspace{-0.5cm}
\end{figure}

\subsection{Fine-tuning vs.~training from scratch}
\label{sec:scratch}
To show the advantage of fine-tuning in the zero-shot scenario, we compare our approach with training from scratch on a single \bench{} class.  
We use our fine-tuning training objective to train ViT encoder from scratch. 
Features from the 6th layer are used. 
Therefore, we keep only the network part up to the 6th layer including. 
Since we train from scratch, we set a larger starting learning rate of $10^{-5}$ for the Adam optimizer.

\cref{fig:from_scratch} shows that training from scratch is prone to overfitting: It results in a drop in Top-1 retrieval accuracy on the test set of the \bench{} class starting from the 150th epoch. 
After the 80th epoch, the accuracy improves very slowly for the \lamp{} class and does not improve anymore for the \chair{} class. 
Note that during training the contrastive loss \cref{eq:contrastive_loss} decreases over all 500 epochs.
Moreover, for all three considered classes, the retrieval accuracy is quite low: it is below $30\%$, while the Top-1 retrieval accuracy of our approach surpasses $70\%$. 
The overfitting result is similar to observations in \cite{chen2023spatially} when they use only one branch to represent both sketch and image modalities.

\subsubsection{3D shape representation: NPR or anime}
\label{sec:representation}
As we have two synthetic sketch styles (\cref{sec:synhteic_styles}), we evaluate our choice of representing 3D shapes with NPR views against views in the anime style. 
\cref{tab:NPR_or_anime} shows a clear advantage of representing 3D shapes using NPR renderings in both considered cases: when query sketches are freehand sketches or synthetic sketches.

\subsubsection{Feature aggregation strategy}
\label{sec:min_vs_avg}
We evaluate our similarity computation strategy  between a query sketch and 3D shape,  given by \cref{eq:zero_shot}, against an alternative strategy of computing the cosine similarity between the query sketch embedding and the average of 3D shape views embeddings:
\begin{equation}
    \mathrm{sim}({Q},{G}) = \mathrm{d}\left(\mathrm{E_\ell}({Q}), \frac{1}{V}\sum_{v \in views}\mathrm{E_\ell}({G_v})\right),
    \label{eq:zero_shot_v2}
\end{equation}
where, as in \cref{eq:zero_shot}, $Q$ and $G$ denote a query sketch and a gallery shape; $G_v$ is a 3D shape view, $V$ is the number of views for an object (5 in our case), $E_\ell(\cdot)$ denotes $\ell$-th layer features of the encoder $E$, and $d$ stands for the cosine similarity.

\begin{table}[ht]
\vspace{-0.1cm}
\centering 
\resizebox{1.0\columnwidth}{!}{
\scriptsize{
\begin{tabular}{l|crrr}
        & \multicolumn{2}{c|}{Chair}                                               & \multicolumn{2}{c}{Lamp}                              \\
        & \multicolumn{1}{l}{acc@1}          & \multicolumn{1}{l|}{acc@5}          & \multicolumn{1}{l}{acc@1} & \multicolumn{1}{l}{acc@5} \\ \hline
Avg. - ViT-CLIP L-6  & \multicolumn{1}{r}{70.32}          & \multicolumn{1}{r|}{\textbf{89.72}}          & 63.06                     & 78.08                     \\
Max. - ViT-CLIP L-6 & \multicolumn{1}{r}{\textbf{74.93}} & \multicolumn{1}{r|}{{89.39}} & \textbf{73.27}            & \textbf{89.49}            \\\hline
Avg. - ViT-CLIP* L-6 & \multicolumn{1}{r}{74.72}          & \multicolumn{1}{r|}{90.71}          &   66.97                   &  82.28                    \\ 
Max. - ViT-CLIP* L-6 & \multicolumn{1}{r}{\textbf{77.11}} & \multicolumn{1}{r|}{\textbf{92.32}} & \textbf{78.38}            & \textbf{92.39}            \\\hline
\end{tabular}%
}}
\vspace{0.05cm} \caption{Comparison of feature selection strategies on the test set of the freehand sketch dataset.} 
\vspace{-0.2cm}
\label{tab:min_vs_avg}
\end{table}

\cref{tab:min_vs_avg} shows the comparison of the two similarity computations strategies for the ViT encoder trained with the CLIP model in zero-shot or our fine-tuned setting. 
It can be seen that in all settings our strategy is superior to this alternative strategy, with a gap of almost $3$ points in Top-1 retrieval accuracy on \emph{chairs}, and of more than $10$ points in both Top-1 and Top-5 on \emph{lamps}.

\subsubsection{Fine-tuning strategies}
We compare our fine-tuning strategy with two alternative strategies of fine-tuning only the weights of layer normalization layers \cite{frankle2020training} and Visual Prompt Tuning (VPT) \cite{jia2022visual}, which we refer to as \emph{ViT-CLIP LayerNorm} and \emph{ViT-CLIP VPT}, respectively.
We train the two additional strategies under the same conditions and loss as our fine-tuning strategy but set a higher learning rate of $10^{-5}$.

The VPT approach consists in adding learnable tokens to the attention layers of the feature extractor. 
During training, all the original network weights are fixed and only the new tokens are updated. 
We use the deep prompt setting and add 5 additional tokens on the first 6 layers of ViT.
As we observe that with VPT the performance on the validation set of the freehand sketch dataset starts to decrease after 100 epochs, we stop the training at 100th epoch and use the last checkpoint.

\begin{table}[ht]
\vspace{-0.2cm}
\centering 
\resizebox{1.0\columnwidth}{!}{
\scriptsize{
\begin{tabular}{l|crrr}
        & \multicolumn{2}{c|}{Chair}                                               & \multicolumn{2}{c}{Lamp}                              \\
        & \multicolumn{1}{l}{acc@1}          & \multicolumn{1}{l|}{acc@5}          & \multicolumn{1}{l}{acc@1} & \multicolumn{1}{l}{acc@5} \\ \hline
ViT-CLIP L-6  & \multicolumn{1}{r}{74.93}          & \multicolumn{1}{r|}{89.39}          & 73.27                     & 89.49                     \\ \hline       
ViT-CLIP LayerNorm L-6  & \multicolumn{1}{r}{74.96}          & \multicolumn{1}{r|}{90.71}          & 73.87                     & 91.59                     \\
ViT-CLIP VPT L-6 & \multicolumn{1}{r}{73.80} & \multicolumn{1}{r|}{90.22} & 73.57          & 90.99            \\
ViT-CLIP* L-6 (Ours) & \multicolumn{1}{r}{\textbf{77.17}}          & \multicolumn{1}{r|}{\textbf{92.32}}          &\textbf{78.38}                     & \textbf{92.39} \\ \hline
\end{tabular}%
}}
\vspace{0.05cm} \caption{Comparison with the alternative fine-tuning strategies on the test set of the dataset of freehand sketches.} 
\label{tab:fine_tuning_strategies}
\vspace{-0.3cm}
\end{table}

\cref{tab:fine_tuning_strategies} shows that both, the layer normalization layer tuning (ViT CLIP LayerNorm L-6) and VPT (ViT CLIP VPT L-6), allow for increased performance compared to the zero-shot ViT (ViT-CLIP L-6) without fine-tuning. However, our fine-tuning strategy (ViT-CLIP* L-6) achieves the best performance.  

\section{Limitations and Future Work}
% \subsection{Best viewpoints for human sketches}
% Our fine-tuning approach and the evaluation gallery consider a given set of views. In our case, this set corresponds to the most likely human sketches viewpoints, as identified by pilot studies of the sketch literature (see Section 4.2). As we use a set of pre-defined viewpoints for training and the gallery, even if this set of views fit well most of the cases, depending on the objects we want to retrieve, it could be possible to improve performance adjusting these viewpoints. For example, airplains are sometimes drawn using a top view, that we do not consider in our experiments. We leave the study of a system that automatically manages best views to future work. 

While the ViT transformer was proven to be a very efficient encoder for an image domain, it might be not the best for sparse sketches. In the case of sketches, non-overlapping patches can contain too little meaningful information and alternative encoder designs should be considered. One such design was recently proposed by Lin et al.~\cite{lin2023zero}. Another direction to explore is to combine vector and raster sketch encoders. To achieve zero-shot performance, the models with tailored encoders then can be trained in a multi-modal setting. 

Next, our fine-tuning strategy can be expanded to include multi-modal training. For example, if textual descriptions of 3D shapes are available, they can be seamlessly integrated into our fine-tuning process.

% \subsection{Multimodal input}
% Our fine-tune features are specialized on the sketch domain and are trained using only images features of a specific layer of the encoder. When using the CLIP encoder, our visual fine-tuned features may not be compared or may not work well when compared with language features. As we think that the combination of text and sketches represent a strong input for different kind of applications, due to their ability to represent well complementary information (e.g. position and details for sketches and style for text), a future work direction could extend our fine-tuning strategy to a multimodal setting. 

\section{Conclusion}
\label{sec:consclusion}
In this work, we introduced an effective zero-shot sketch-based 3D shape retrieval method. 
We demonstrated how to efficiently adapt models pretrained on different pretext tasks, like CLIP, to the studied problem. 
We show that it is possible to fine-tune a model leveraging only synthetic sketches of a single shape category and demonstrated generalization to freehand abstract sketches of other shape categories. We also showed that performance is similar independently of the choice of a shape category for fine-tuning. 
We bring insights into the role of object scale in the image plane and provide recommendations taking into account query abstraction. We compare the performance of two popular image encoders ViT and ResNet and show that the same object scale is beneficial for the two encoders under consideration regardless of the pretext task used.
We also carefully study the role of object scale in the image plane and provide recommendations taking into account query abstraction. 
We believe that our work provides valuable information for methods aimed at assessing the perceptual similarity between sketches in different styles.

{\small
\bibliographystyle{ieee_fullname}
\bibliography{egbib}
}

\clearpage
\appendix
\begin{figure*}[t]
% \vspace{1cm}
\includegraphics[width=\linewidth]{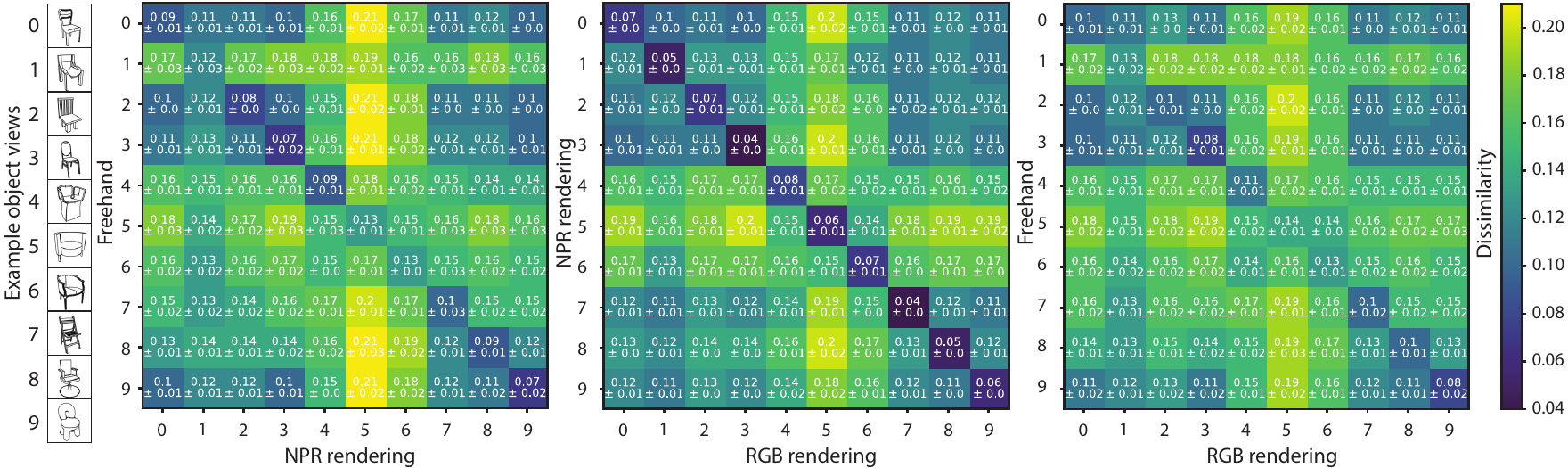}
% \label{fig:objects_similarity}
\vspace{-0.5cm}
\caption{We plot pairwise distances between shapes, when their views come from one of the three image domains: freehand sketches \cite{qi2021toward}, line or RGB renderings. See \cref{sec:Analysis} for the details.}
\label{fig:ten_objects}
\end{figure*}

\section{How well can CLIP differentiate between views and objects?}
\label{sec:Analysis}
% \todo{the study we did for CVMP. This section is not necessary, but might be a nice addition to the paper}
% \todo{Let's see later if to keep this section. Maybe move to the supplemental?}

Our work targets retrieval from freehand quick and abstract sketches, such as sketches collected by Qi et al.~\cite{qi2021toward} and Zhang et al.~\cite{zhang2021sketch2model}. 
There are several strategies to represent 3D shapes using their multi-view projections in sketch-based 3D model retrieval literature: using RGB renderings or NPR renderings. 
However, there is a large domain gap between such views and freehand sketches we consider as queries.
To motivate our work, we study CLIP \cite{clip} features similarity across domains, and its ability to match the model views across various domains.

In this section, we only consider 3D shapes from the chair category from the \emph{AmateurSketch} dataset \cite{qi2021toward}.
\begin{wrapfigure}{r}{0.3\linewidth}
\vspace{-0.4cm}
\hspace{-0.8cm}
    \centering
    \includegraphics[height=3.2cm]{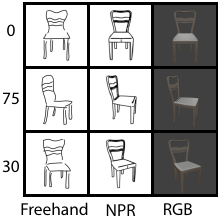}
\vspace{-0.2cm}
\end{wrapfigure}
We use three different views for every object, with the camera azimuth angles set to $0^\circ$, $30^\circ$ and $75^\circ$.
The inset on the right shows example viewpoints for one of the considered 3D shapes.
In this section, we use the third layer of the pre-trained ResNet101 CLIP image encoder as our CLIP embedding space, following \cite{CLIPasso}.
% \todo{the little figure has "Synth." as name while in the paper I think we will refer to it as "line rendering"}

\paragraph{Aggregated feature similarity} We consider $10$ randomly-selected objects (the corresponding freehand sketches for them are shown in \cref{fig:ten_objects}), and compute the distance between two 3D shapes $A$ and $B$, represented with their multi-view renderings or freehand sketches from the three considered viewpoints, as follows:
\begin{equation}
d(A^i,B^j) = \frac{1}{V} \sum_{k=1}^{V}\left(\mathrm{CLIP}(A_{k}^i) - \mathrm{CLIP}(B_{k}^j)\right)^2
\end{equation}
where $V = 3$ is the number of views, and subscripts $i$ and $j$ denote one of three image domains: freehand sketches, NPR renderings or RGB renderings.

In \cref{fig:ten_objects}, we plot pairwise distances between shapes, when their views come from one of the three image domains.  
We can see that in all three configurations, comparing the same object between different domains results in the lowest average distance (darker color) in most of the cases.
This shows general robustness of the CLIP model across different domains. 
We also can see that it is easier to find correct matches for 3D shapes represented with freehand sketches and NPR renderings than for 3D shapes represented with freehand sketches and RGB renderings.
% However, failure cases are possible, in particular, when amateur sketches are concerned: Sometimes the loss in CLIP embedding space can not confidently distinguish two shapes in two different domains. 
% Sometimes comparing the same object does not result in the lowest distance or there is overlap between confidence intervals. 

\paragraph{Matching views across various domains}
\begin{figure}[h!]
    \centering
    \includegraphics[width=\linewidth]{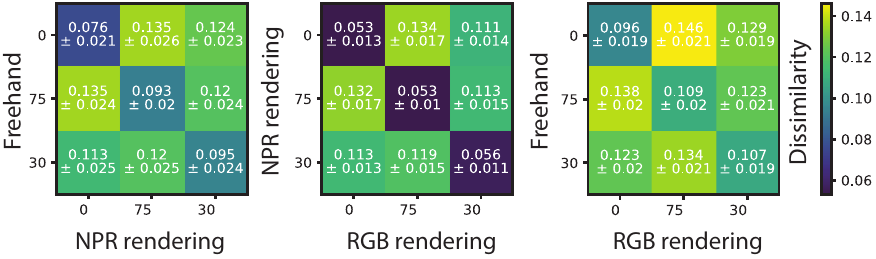}
    \vspace{-0.5cm}
    \caption{We plot average pairwise distances between views in different image domains: freehand sketches \cite{qi2021toward}, NPR or RGB renderings. See \cref{sec:Analysis} for the details.}
    \label{fig:views}
\end{figure}
Comparing individual views $k$ and $h$ in domains $i$ and $j$, we compare view embeddings of the same object for the compared domains, and average over different objects:
\begin{equation}
d(k^i,h^j) = \frac{1}{N} \sum_{A=\mathrm{obj}_1}^{\mathrm{obj}_N}\left(\mathrm{CLIP}(A^i_{k}) - \mathrm{CLIP}(A_{h}^j)\right)^2
\end{equation}
where $N=100$ is the number of objects. 
\cref{fig:views} shows that for all configurations the average lowest values are obtained when comparing the same view.
This shows that, in general, CLIP is able to match the same view of an object between different domains. 
Similarly, we observe that matching views of freehand sketches with NPR renderings is more reliable than matches of freehand sketches with RGB renderings.
However, when comparing freehand sketch views with views from other domains, we observe that the confidence intervals can overlap. This means, that occasionally an incorrect viewpoint in a different domain can be selected.

\end{document}